\documentclass{IEEEtran}
\usepackage[utf8]{inputenc}
\usepackage{graphicx}
\usepackage{epstopdf}
\usepackage{amsmath,amssymb,booktabs,tabularx}
\usepackage[hyphens]{url}
\usepackage[pagebackref=true,breaklinks=true]{hyperref}
\hypersetup{
    colorlinks,
    linkcolor={red!75!black},
    citecolor={blue!75!black},
    urlcolor={blue!75!black},
}
\usepackage[capitalise]{cleveref}

\usepackage{mathrsfs}
\usepackage{float}
\usepackage{multirow}
\usepackage{soul}
\usepackage[table]{xcolor}
\usepackage{comment}
\usepackage{subcaption}
\usepackage[super]{nth}
\usepackage{mwe}
\usepackage{slashbox}
\usepackage{authblk}
\usepackage{multicol}
\usepackage{import}
\usepackage{siunitx}
\usepackage{mathtools}
\usepackage{cite}
\usepackage{microtype}

\usepackage{makecell}
\usepackage[inline]{enumitem}
\setlist*[enumerate]{label=(\arabic*)}

\usepackage{tikz}
\usepackage{pgfplots}
\usetikzlibrary{spy,calc}
\usepackage{xspace}
\makeatletter
\DeclareRobustCommand\onedot{\futurelet\@let@token\@onedot}
\newcommand{\@onedot}{\ifx\@let@token.\else.\null\fi\xspace}
\makeatother

\newcommand{\cf}{cf\onedot}

\newcommand{\etal}[1]{#1 \emph{et~al\onedot}}
\newcommand{\ie}{i.\,e.,\xspace}
\newcommand{\eg}{e.\,g.,\xspace}

\newif\ifblackandwhitecycle
\gdef\patternnumber{0}

\pgfkeys{/tikz/.cd,
	zoombox paths/.style={
		draw=orange,
		very thick
	},
	black and white/.is choice,
	black and white/.default=static,
	black and white/static/.style={ 
		draw=white,   
		zoombox paths/.append style={
			draw=white,
			postaction={
				draw=black,
				loosely dashed
			}
		}
	},
	black and white/static/.code={
		\gdef\patternnumber{1}
	},
	black and white/cycle/.code={
		\blackandwhitecycletrue
		\gdef\patternnumber{1}
	},
	black and white pattern/.is choice,
	black and white pattern/0/.style={},
	black and white pattern/1/.style={    
		draw=white,
		postaction={
			draw=black,
			dash pattern=on 2pt off 2pt
		}
	},
	black and white pattern/2/.style={    
		draw=white,
		postaction={
			draw=black,
			dash pattern=on 4pt off 4pt
		}
	},
	black and white pattern/3/.style={    
		draw=white,
		postaction={
			draw=black,
			dash pattern=on 4pt off 4pt on 1pt off 4pt
		}
	},
	black and white pattern/4/.style={    
		draw=white,
		postaction={
			draw=black,
			dash pattern=on 4pt off 2pt on 2 pt off 2pt on 2 pt off 2pt
		}
	},
	zoomboxarray inner gap/.initial=5pt,
	zoomboxarray columns/.initial=2,
	zoomboxarray rows/.initial=2,
	subfigurename/.initial={},
	figurename/.initial={zoombox},
	zoomboxarray/.style={
		execute at begin picture={
			\begin{scope}[
				spy using outlines={%
					zoombox paths,
					width=\imagewidth / \pgfkeysvalueof{/tikz/zoomboxarray columns} - (\pgfkeysvalueof{/tikz/zoomboxarray columns} - 1) / \pgfkeysvalueof{/tikz/zoomboxarray columns} * \pgfkeysvalueof{/tikz/zoomboxarray inner gap} -\pgflinewidth,
					height=\imageheight / \pgfkeysvalueof{/tikz/zoomboxarray rows} - (\pgfkeysvalueof{/tikz/zoomboxarray rows} - 1) / \pgfkeysvalueof{/tikz/zoomboxarray rows} * \pgfkeysvalueof{/tikz/zoomboxarray inner gap}-\pgflinewidth,
					magnification=3,
					every spy on node/.style={
						zoombox paths
					},
					every spy in node/.style={
						zoombox paths
					}
				}
				]
			},
			execute at end picture={
			\end{scope}
			\node at (image.north) [anchor=north,inner sep=0pt] {\subcaptionbox{\label{\pgfkeysvalueof{/tikz/figurename}-image}}{\phantomimage}};
			\node at (zoomboxes container.north) [anchor=north,inner sep=0pt] {\subcaptionbox{\label{\pgfkeysvalueof{/tikz/figurename}-zoom}}{\phantomimage}};
			\gdef\patternnumber{0}
		},
		spymargin/.initial=0.5em,
		zoomboxes xshift/.initial=1,
		zoomboxes right/.code=\pgfkeys{/tikz/zoomboxes xshift=1},
		zoomboxes left/.code=\pgfkeys{/tikz/zoomboxes xshift=-1},
		zoomboxes yshift/.initial=0,
		zoomboxes above/.code={
			\pgfkeys{/tikz/zoomboxes yshift=1},
			\pgfkeys{/tikz/zoomboxes xshift=0}
		},
		zoomboxes below/.code={
			\pgfkeys{/tikz/zoomboxes yshift=-1},
			\pgfkeys{/tikz/zoomboxes xshift=0}
		},
		caption margin/.initial=4ex,
	},
	adjust caption spacing/.code={},
	image container/.style={
		inner sep=0pt,
		at=(image.north),
		anchor=north,
		adjust caption spacing
	},
	zoomboxes container/.style={
		inner sep=0pt,
		at=(image.north),
		anchor=north,
		name=zoomboxes container,
		xshift=\pgfkeysvalueof{/tikz/zoomboxes xshift}*(\imagewidth+\pgfkeysvalueof{/tikz/spymargin}),
		yshift=\pgfkeysvalueof{/tikz/zoomboxes yshift}*(\imageheight+\pgfkeysvalueof{/tikz/spymargin}+\pgfkeysvalueof{/tikz/caption margin}),
		adjust caption spacing
	},
	calculate dimensions/.code={
		\pgfpointdiff{\pgfpointanchor{image}{south west} }{\pgfpointanchor{image}{north east} }
		\pgfgetlastxy{\imagewidth}{\imageheight}
		\global\let\imagewidth=\imagewidth
		\global\let\imageheight=\imageheight
		\gdef\columncount{1}
		\gdef\rowcount{1}
		
	},
	image node/.style={
		inner sep=0pt,
		name=image,
		anchor=south west,
		append after command={
			[calculate dimensions]
			node [image container,subfigurename=\pgfkeysvalueof{/tikz/figurename}-image] {\phantomimage}
			node [zoomboxes container,subfigurename=\pgfkeysvalueof{/tikz/figurename}-zoom] {\phantomimage}
		}
	},
	color code/.style={
		zoombox paths/.append style={draw=#1}
	},
	connect zoomboxes/.style={
		spy connection path={\draw[draw=none,zoombox paths] (tikzspyonnode) -- (tikzspyinnode);}
	},
	help grid code/.code={
		\begin{scope}[
			x={(image.south east)},
			y={(image.north west)},
			font=\footnotesize,
			help lines,
			overlay
			]
			\foreach \x in {0,1,...,9} { 
				\draw(\x/10,0) -- (\x/10,1);
				\node [anchor=north] at (\x/10,0) {0.\x};
			}
			\foreach \y in {0,1,...,9} {
				\draw(0,\y/10) -- (1,\y/10);                        \node [anchor=east] at (0,\y/10) {0.\y};
			}
		\end{scope}    
	},
	help grid/.style={
		append after command={
			[help grid code]
		}
	},
}

\newcommand\phantomimage{%
	\phantom{%
		\rule{\imagewidth}{\imageheight}%
	}%
}
\newcommand\zoombox[2][]{
	\begin{scope}[zoombox paths]
		\pgfmathsetmacro\xpos{
			(\columncount-1)*(\imagewidth / \pgfkeysvalueof{/tikz/zoomboxarray columns} + \pgfkeysvalueof{/tikz/zoomboxarray inner gap} / \pgfkeysvalueof{/tikz/zoomboxarray columns} ) + \pgflinewidth
		}
		\pgfmathsetmacro\ypos{
			(\rowcount-1)*( \imageheight / \pgfkeysvalueof{/tikz/zoomboxarray rows} + \pgfkeysvalueof{/tikz/zoomboxarray inner gap} / \pgfkeysvalueof{/tikz/zoomboxarray rows} ) + 0.5*\pgflinewidth
		}
		\edef\dospy{\noexpand\spy [
			#1,
			zoombox paths/.append style={
				black and white pattern=\patternnumber
			},
			every spy on node/.append style={#1},
			x=\imagewidth,
			y=\imageheight
			] on (#2) in node [anchor=north west] at ($(zoomboxes container.north west)+(\xpos pt,-\ypos pt)$);}
		\dospy
		\pgfmathtruncatemacro\pgfmathresult{ifthenelse(\columncount==\pgfkeysvalueof{/tikz/zoomboxarray columns},\rowcount+1,\rowcount)}
		\global\let\rowcount=\pgfmathresult
		\pgfmathtruncatemacro\pgfmathresult{ifthenelse(\columncount==\pgfkeysvalueof{/tikz/zoomboxarray columns},1,\columncount+1)}
		\global\let\columncount=\pgfmathresult
		\ifblackandwhitecycle
		\pgfmathtruncatemacro{\newpatternnumber}{\patternnumber+1}
		\global\edef\patternnumber{\newpatternnumber}
		\fi
	\end{scope}
}

\makeatletter
\let\old@ps@headings\ps@headings
\let\old@ps@IEEEtitlepagestyle\ps@IEEEtitlepagestyle
\def\confheader#1{%
	\def\ps@IEEEtitlepagestyle{%
		\old@ps@IEEEtitlepagestyle%
		\def\@oddhead{\strut\hfill#1\hfill\strut}%
		\def\@evenhead{\strut\hfill#1\hfill\strut}%
	}%
	\ps@headings%
}
\makeatother

\confheader{%
	\parbox{20cm}{SUBMITTED TO IEEE TRANSACTIONS ON GEOSCIENCE AND REMOTE SENSING FOR POSSIBLE PUBLICATION.}
}

\title{Pixel-wise Distance Regression for Glacier Calving Front Detection and Segmentation}

\author{Amirabbas Davari*, Christoph Baller*, Thorsten Seehaus, Matthias Braun, Andreas Maier, Vincent Christlein
	\thanks{A.\ Davari, C.\ Baller, A.\ Maier and V.\ Christlein are with the Computer Science Department at Friedrich-Alexander University Erlangen-Nürnberg, 91058 Erlangen, Germany (email: amir.davari@fau.de).}
	\thanks{T.\ Seehaus and M.\ Braun are with the Geography and Geosciences Department at Friedrich-Alexander University Erlangen-Nürnberg, 91058 Erlangen, Germany.}
	\thanks{* Amirabbas Davari and Christoph Baller contributed equally to this work.}
	\thanks{Grateful acknowledgment is made to the German Aerospace Center (DLR), the European Space Agency (ESA), and the Alaska Satellite Facility (ASF) for providing the SAR data for our analysis.}}

\IEEEtitleabstractindextext{%
	\begin{abstract}
Glacier calving front position (CFP) is an important glaciological variable. Traditionally, delineating the CFPs has been carried out manually, which was subjective, tedious and expensive. Automating this process is crucial for continuously monitoring the evolution and status of glaciers. Recently, deep learning approaches have been investigated for this application. However, the current methods get challenged by a severe class-imbalance problem. In this work, we propose to mitigate the class-imbalance between the calving front class and the non-calving front class by reformulating the segmentation problem into a pixel-wise regression task. A Convolutional Neural Network gets optimized to predict the distance values to the glacier front for each pixel in the image. The resulting distance map localizes the CFP and is further post-processed to extract the calving front line. We propose three post-processing methods, one method based on statistical thresholding, a second method based on conditional random fields (CRF), and finally the use of a second U-Net. The experimental results confirm that our approach significantly outperforms the state-of-the-art methods and produces accurate delineation. The Second U-Net obtains the best performance results, resulting in an average improvement of about \SI{21}{\percent} dice coefficient enhancement. 
	\end{abstract}
	
	\begin{IEEEkeywords}
		distance map regression, glacier calving front segmentation, SAR imagery, U-Net, class-imbalance
	\end{IEEEkeywords}
}

\begin{document}

\maketitle
\IEEEdisplaynontitleabstractindextext

\section{Introduction}\label{sec:introduction}
Many glaciers are draining directly into the ocean or lakes (calving glaciers). In particular along the Antarctic Peninsula, in Patagonia or along Greenlands margins, the ice bodies are discharging ice via outlet glaciers, which are typically located in narrow fjords. The ice front position of calving glaciers can vary significantly and impact the ice dynamics. The retreat of the ice front can reduce the lateral buttressing forces or detach the ice from pinning points (bedrock) and thus destabilize the ice flow and cause further ice front recession~\cite{furst2016safety}. 
For example, the retreat and disintegration of ice shelves along the Antarctic Peninsula has led to a sharp increase in ice discharge from tributary glaciers, causing an acceleration of glacier thinning, further retreat, and consequently high ice mass losses~\cite{friedl2018recent, scambos2004glacier}. 

Changes in the ice front position can also be linked to variations in the grounding zone (transition zone where a calving glacier starts to float)~\cite{friedl2020remote}. 
Deeper bedrock elevations inland from the grounding zone (retrograde bedrock) will lead to a feedback mechanism called marine ice-sheet instability as the grounding zone retreats, resulting in further grounding zone retreat and increasing ice loss~\cite{robel2016persistence}.
Therefore, monitoring of glacier front positions can serve as an indicator for changes in ice dynamics and provides crucial information for analyzing the ongoing processes, \eg for model initialization or calibration.

Optical and SAR remote sensing imagery are typically used to map the position of the glacier front~\cite{baumhoer2018remote}. 
However, cloud cover and polar night limit the coverage by optical imagery, which is not the case for SAR data. Up to now, visual inspection and manual mapping of the glacier front is commonly used. In polar regions, the water in front of the glacier front is frequently covered by the so-called ice-mélange, consisting of sea ice and icebergs, which can look quite similar to a glacier surface on remote sensing imagery. Thus, the analyst needs to carefully separate between the ice-mélange and the actual glacier, making the analysis difficult and laborious. Moreover, such human analyses are always subjective and can vary from analyst to analyst~\cite{paul2013accuracy}. 
Several semi-automatic and automatic ice front mapping techniques were developed, which typically rely on edge enhancement, image classification or edge detection. \etal{Baumhoer}~\cite{baumhoer2018remote} provide a detailed review of the various methods on this topic.

Recently, the first studies on using Convolutional Neural Networks (CNN) for glacier front mapping appeared. \etal{Mohajerani}~\cite{mohajerani2019detection} trained and tested a U-Net architecture on multi-spectral Landsat 5, 7, and 8 acquisitions at four outlet glaciers of the Greenland ice sheet. In a similar way, \etal{Baumhoer}~\cite{baumhoer2019automated} and \etal{Zhang}~\cite{zhang2019automatically} mapped the ice fronts on SAR imagery from Sentinel-1 and TerraSAR-X.
Both studies classify the surface region pixel-wise into the two types, \ie ice mélange and non-ice mélange, and extract the final calving front position (CFP) in a post-processing step using edge detection algorithms. 

A direct prediction of the glacier CFP is very challenging as the actual front line that the model should learn to predict contributes a very small portion to the number of image pixels. This leads to a severe class-imbalance, which is known to be an Achilles heel for supervised learning algorithms. While predicting the calving front lines directly, \etal{Mohajerani}~\cite{mohajerani2019detection} tackled the  class-imbalance problem using a custom sample weight that penalizes miss-classified pixels of the calving front higher. 
In the medical image processing domain, applications like segmenting the optic disc (OD) and fovea in retinal images struggle with the class-imbalance problem, too. \etal{Meyer}~\cite{meyer2018pixel} proposed to tackle this problem by reformulating the segmentation task to a pixel-wise regression that jointly detects OD and fovea location in a retinal image. The method achieved high detection results close to the human observer performance.

In this paper, we propose a strategy for glacier front segmentation using a modified U-Net with a similar architecture used by \etal{Zhang}~\cite{zhang2019automatically}, which is adjusted for semantic segmentation of SAR imagery. Getting inspired by \etal{Meyer}~\cite{meyer2018pixel}, we first reformulate the segmentation problem into a pixel-wise regression task and train the U-Net to predict the distance map transform of the glacier fronts in the SAR images. The output of this step is not only accurate and robust against the class-imbalance, but also emphasizes the region of interest, \ie the glacier front line and can serve for detection purposes and human annotation. In the next step, in an attempt to automatically extract the actual glacier front locations from the predicted distance maps, we propose the use of three distinct approaches, namely statistical threshold, a conditional random field, and a second U-Net. The rest of the presentation in this paper is organized as follows: \cref{sec:methodology} reviews the tools used in this work and explains the proposed algorithms. \Cref{sec:evaluation} is dedicated to demonstrating the dataset, explaining the evaluation setup, and presenting and discussing the quantitative and qualitative results. Finally, \cref{sec:conclusion} concludes the paper and discusses the outlook.

\section{Methodology}\label{sec:methodology}
\begin{figure}
	\centering
	\includegraphics[width=1\linewidth]{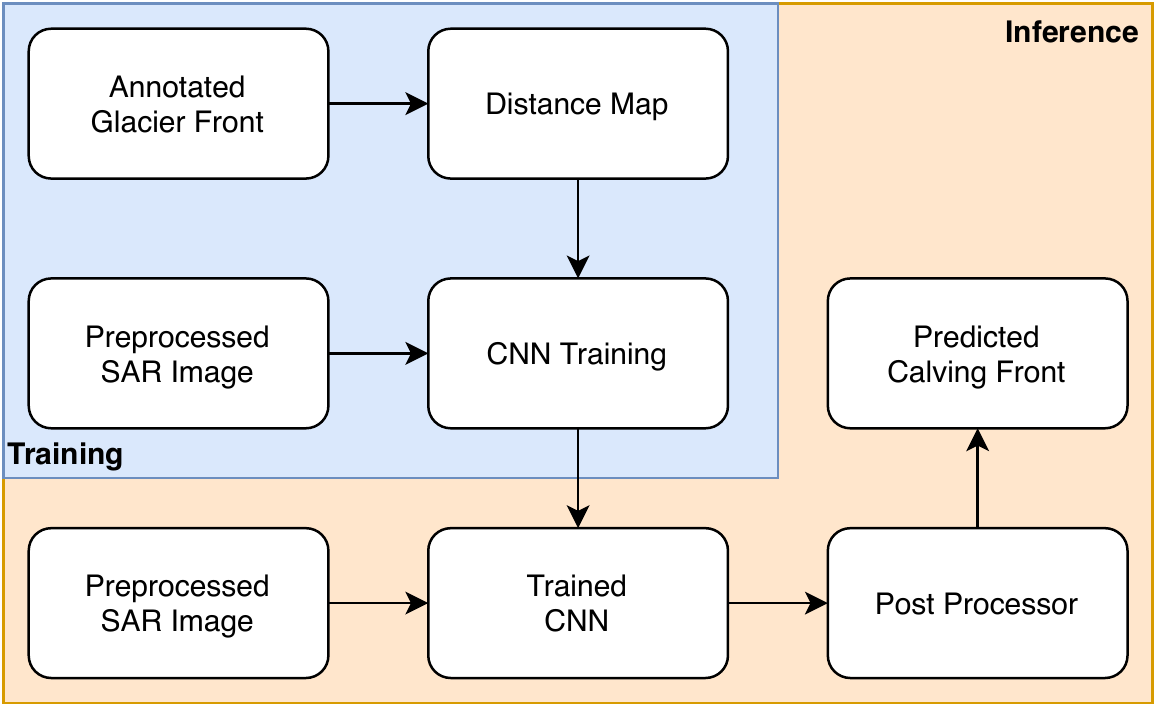}
	\caption{Outline of our proposed workflow for automated calving front position segmentation.}
	\label{fig:overview}
\end{figure}

\Cref{fig:overview} depicts the overview of our proposed workflow. It consists of three key processing blocks:\ the distance map transform, CNN training, and post-processing. This section is dedicated to explaining these blocks in detail.
\subsection{Distance Map Transform}\label{sec:method_DistMapTransform}
Glacier front detection can be described as a semantic image segmentation problem that is usually resolved by a binary classification, where the classes are determined pixel-wise by their affiliation to the glacier front.
The problem is the high imbalance between the frequencies of the glacier front class and non-glacier front class samples, which significantly challenges the optimization of the model.
To mitigate this issue, we propose to follow \etal{Meyer}~\cite{meyer2018pixel} and reformulate the semantic segmentation problem at hand into a regression task, where the model approximates the closeness of each pixel in the image to the glacier front line.

\begin{figure}
	\begin{subfigure}[b]{0.32\linewidth}
		\centering
		\includegraphics[width=1\linewidth]{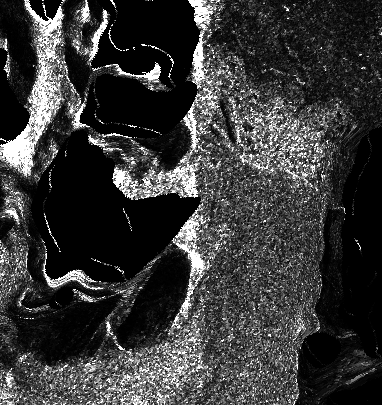}
		\caption{}\label{fig:dist_map_a}
	\end{subfigure}
	\begin{subfigure}[b]{0.32\linewidth}
		\centering
		\includegraphics[width=1\linewidth]{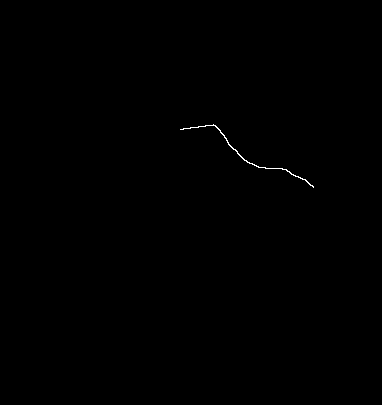}
		\caption{}\label{fig:dist_map_b}
	\end{subfigure}
	\begin{subfigure}[b]{0.32\linewidth}
		\centering
		\includegraphics[width=1\linewidth]{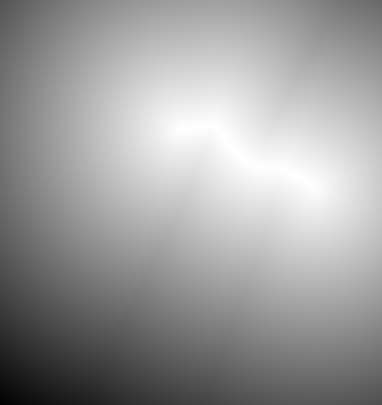}
		\caption{}\label{fig:dist_map_c}
	\end{subfigure}
	
	\vspace{0.5em}
	\begin{subfigure}[b]{0.32\linewidth}
		\centering
		\includegraphics[width=1\linewidth]{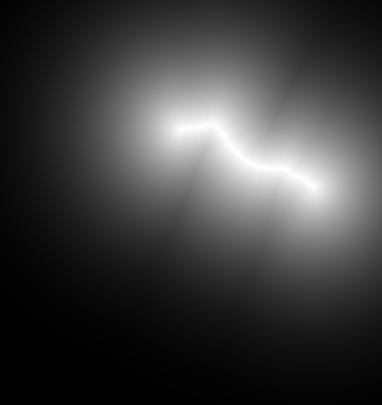}
		\caption{}\label{fig:dist_map_d}
	\end{subfigure}
	\begin{subfigure}[b]{0.32\linewidth}
		\centering
		\includegraphics[width=1\linewidth]{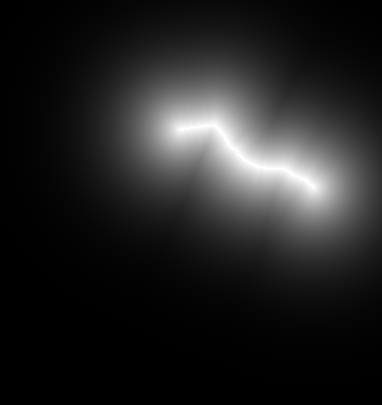}
		\caption{}\label{fig:dist_map_e}
	\end{subfigure}
	\begin{subfigure}[b]{0.32\linewidth}
		\centering
        \includegraphics[width=1\linewidth]{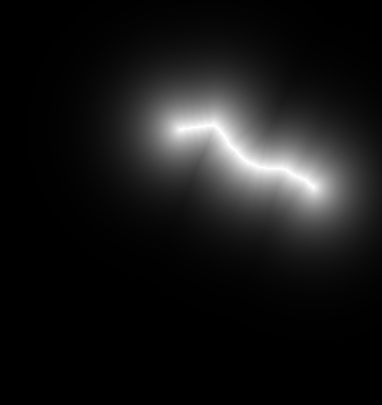}
        \caption{}\label{fig:dist_map_f}
	\end{subfigure}
	\caption{\subref{fig:dist_map_a} Sample SAR image from the Mapple glacier (ENVISAT \mbox{2010-07-03}), \subref{fig:dist_map_b} glacier calving front position ground truth, \subref{fig:dist_map_c}--\subref{fig:dist_map_f} corresponding ground truth distance map with $\gamma \in \{1,4,7,10\}$.}
	\label{fig:dist_map}
\end{figure}

The distance map for each pixel $p(x,y) \in \Omega$ represents the distance to the glacier front, where $\Omega \subset \mathbb{R}^{2}$ denotes the image domain.
For each image in the dataset, we first binarize the ground truth calving front line, invert it so that the glacier front forms the background, and then apply the Euclidean distance transform (EDT).
\begin{equation}\label{eq:edt}
	D(x,y) = \text{EDT}\Bigl(\text{invert}\bigl(p(x,y)\bigr)\Bigr)
\end{equation}
Afterwards, we normalize the distance map $D$, where the pixel intensities decrease relative to their distance from the glacier front location;
\begin{equation}\label{eq:dist_map}
	D_{\text{norm}}(x, y) = \left(1 - \frac{D(x, y)}{\max D(x,y)}\right)^{\gamma}\enspace,
\end{equation}
where $\gamma$ is a decay parameter determining the spread of $D_{\text{norm}}$. The influence of the decay parameter is visualized in Fig. \ref{fig:dist_map}.
%

\subsection{Deep Convolutional Neural Network}\label{sec:cnn}
In order to transform the initial pixel-wise binary classification problem into a regression task, we need to train a model capable of accurately approximating $D_{\text{norm}}(x, y)$ given a Glacier SAR image.
Convolutional Neural Networks (CNN) have revolutionized image analysis and computer vision domains. This work uses a CNN with a U-Net architecture~\cite{DBLP:journals/corr/RonnebergerFB15}, which has successfully been used for biomedical image segmentation. The U-Net consists of a contraction path followed by an expansion path that is connected via skip-connections. Our model is inspired by \etal{Zhang}~\cite{zhang2019automatically}, who adjusted the U-Net architecture for semantic segmentation of SAR imagery. 
The SAR images are the input for the model. The optimization process tries to minimize the mean squared error (MSE) loss between the output of the neural network and the distance map transform of the corresponding glacier front mask of the input.

\subsection{Post processing}\label{sec:post-processing}
The model predicts a smooth distance map, where the pixel intensities increase the closer they are to the glacier front. This map helps to localize the calving front line and narrows down the searching area for the region of interest. In the next step, the actual glacier calving front line should be extracted from the estimated distance map. Computing the maximum pixel intensity results in some sparse and noisy points that do not represent the CFP at all. In this work, we propose three different approaches to approximate the location of the glacier front, namely statistical thresholding, conditional random fields using guided filters, and a second U-Net as a learning-based approach.

\subsubsection{Statistical Thresholding (Threshold)}
\begin{figure}
	\centering
	\begin{subfigure}[b]{0.32\linewidth}
		\centering
        \includegraphics[width=1\linewidth]{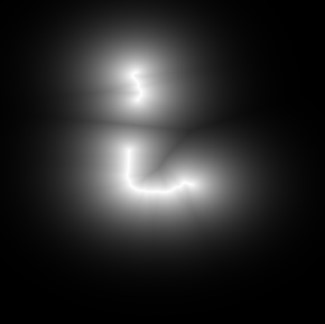}
        \caption{}
        \label{fig:stat_thresholding_a}
	\end{subfigure}
			\medskip
	\begin{subfigure}[b]{0.32\linewidth}
		\centering
		\includegraphics[width=1\linewidth]{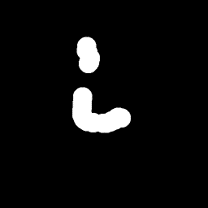}
		\caption{}
        \label{fig:stat_thresholding_b}
	\end{subfigure}
		\medskip
	\begin{subfigure}[b]{0.32\linewidth}
		\centering
        \includegraphics[width=1\linewidth]{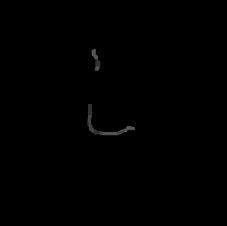}
        \caption{}
        \label{fig:stat_thresholding_c}
	\end{subfigure}
	\caption{Statistical Thresholding post processing steps: \subref{fig:stat_thresholding_a} predicted distance map, \subref{fig:stat_thresholding_b} binarized distance map, and \subref{fig:stat_thresholding_c} calving front line.} 
	\label{fig:stat_thresholding}
\end{figure}

Probably, the most intuitive solution to this problem is to threshold the distance map on a specific value. The prediction gets divided into two classes, one of which includes the glacier front line. Very small threshold values result in many false alarms while too large threshold values result in the exclusion of the correct CFP pixel and hence a severely discrete line. 
Furthermore, the pixel intensity value range in the predicted distance maps marking the CFPs vary for different SAR images. Therefore, the threshold needs to be adaptive and be selected for each image individually. To achieve this, we create a histogram of the distance map, which contains the quantity of all pixel intensities. 
We chose the threshold such that \SI{95}{\percent} of the intensity bins are belonging to the background. This ensures the extraction of a narrow region in the image containing the glacier front and not being polluted by too many false alarms. To further prune this area, we apply morphological thinning operation producing a skeleton that defines the final calving front line. Figure \ref{fig:stat_thresholding} illustrates these steps for a sample test data.

\subsubsection{Conditional Random Field}
Deep Convolutional Neural Networks tend to yield smooth responses. They can predict the presence and rough position of objects during the image segmentation but have difficulties to outline their borders and specifically sharp edges. To mitigate this problem, the location accuracy of a fully connected Conditional Random Field (CRF) in combination with the Neural Network was purposed in the DeepLab system~\cite{DBLP:journals/corr/ChenPK0Y16}. With this approach, we use the same strategy on our predicted distance maps.
The CRF model employs the energy function
\begin{equation}\label{eq:crf_1}
	E(x) = \sum_{i} \theta_{i}(x_{i}) + \sum_{ij} \theta_{ij}(x_{i}, x_{j})
\end{equation}
with the unary potential $\theta_{i}(x_{i}) = - \log P(x_{i})$, where $P(x_{i})$ is the individual label assignment probability of the distance map model, $\theta_{ij}(x_{i}, x_{j})$ is the pairwise potential, and  $x$ is the final label assignment for the pixel. We build the foreground and background classes of $P(x_i)$ by setting the background class probabilities as the inverse of the individual foreground class probability. The pairwise potential consists of a bilateral and a Gaussian kernel
\begin{multline}\label{eq:crf_2}
	\textstyle
	\theta_{ij}(x_{i}, x_{j}) = \mu(x_{i}, x_{j})
	\Bigl[ 
	\omega_{1} \exp\bigl(-\frac{\lVert p_{i}-p_{j} \rVert^{2}}{2\sigma_\alpha^{2}} - \frac{\lVert I_{i}-I_{j} \rVert^{2}}{2\sigma_\beta^{2}}\bigr) \\ +
	\omega_{2} \exp\bigl(-\frac{\lVert p_{i}-p_{j} \rVert^{2}}{2\sigma_\gamma^{2}}\bigr)
	\Bigr] \enspace,
\end{multline}
where $\theta$ denotes the simple label compatibility function that is given by Potts model $\theta_{ij}(x_{i}, x_{j}) = [x_{i}\ne x_{j}]$, which only penalize nodes with distinct labels. The kernel potentials depend on pixel positions $p$ and grayscale intensities $I$. The parameters $\sigma_\alpha$, $\sigma_\beta$, and $\sigma_\gamma$ control the scale of the Gaussian kernels and the linear combination weights $\omega_1$, $\omega_2$ manage the individual influence of the kernel potentials. Thus, the CRF model efficiently approximates the probabilistic inference using Gaussian filtering in feature space~\cite{DBLP:journals/corr/abs-1210-5644}. 
This results in a specific class assignment of the calving front given the probabilities from the predicted distance map and its associated SAR image.

\subsubsection{Second U-Net} 
Convolutional Neural Networks are successful and widely used to tackle today's computer vision problems. As described before, extracting the calving front line from the estimated distance map is challenging for various reasons, which can be tackled within a semantic segmentation regime using a U-Net. The output of the pixel-wise regression model in the first step of the workflow has the same dimension and shape as the original input and can be utilized as the input for a new segmentation model in the post-processing step.

In this approach, we use a second U-Net that takes the previously estimated distance maps and predicts the final glacier front. The neural network shares the same U-Net architecture we introduced in \cref{sec:cnn}. 
In order to train the model, we first reduce the still existing class-imbalance 
by thickening the glacier front lines using morphological dilation, which would otherwise substantially degrade the model optimization capabilities. 
We use the binary cross-entropy (BCE) loss function to calculate the difference between the predicted line and the ground truth glacier front mask.


\begin{figure*}
	\centering
	\includegraphics[width=0.9\linewidth]{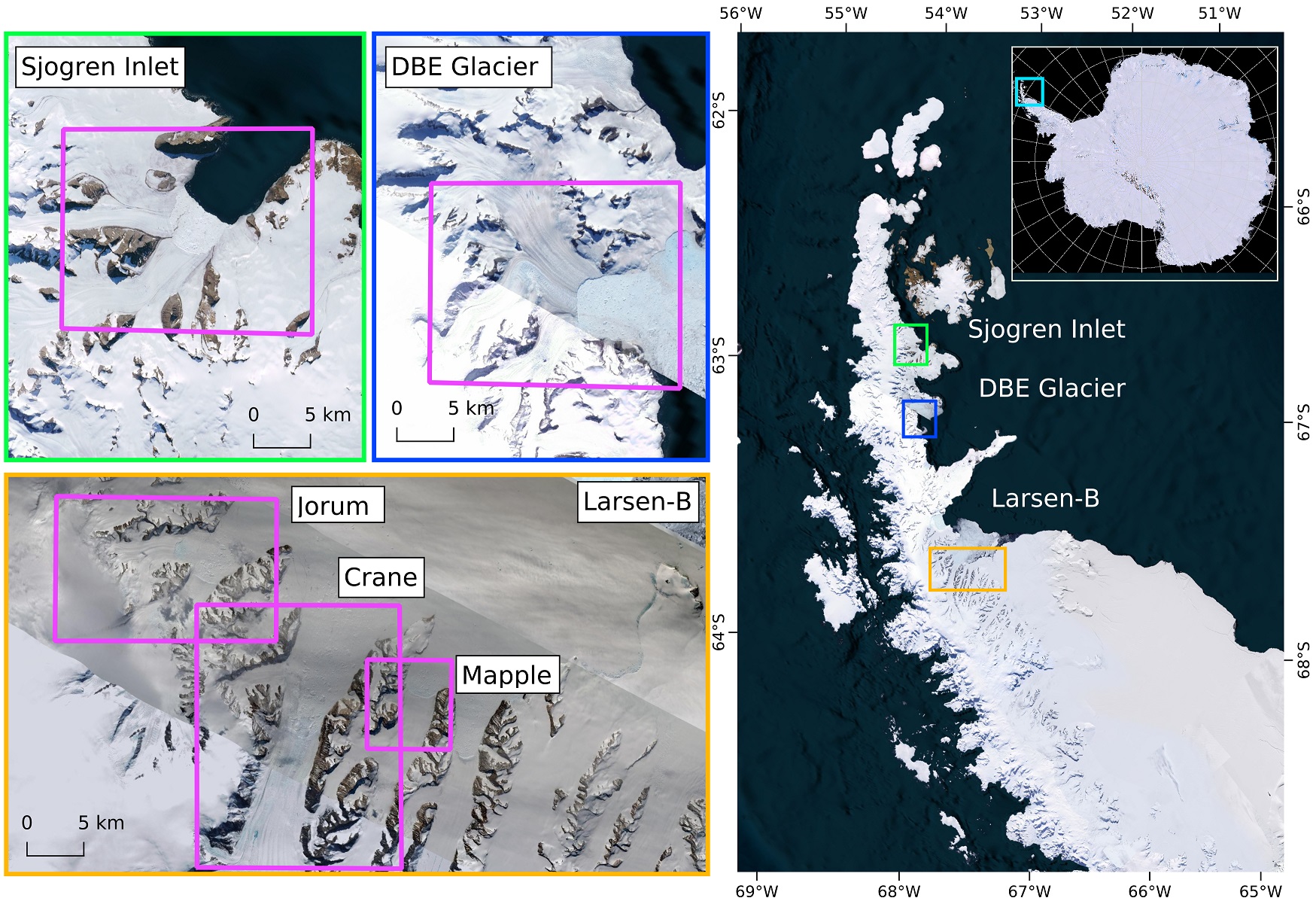}
	\caption{Overview of studied glaciers on the northern Antarctic Peninsula. Pink polygons indicate the subsets used in the analysis. Background: Bing Aerial Maps © Microsoft and Google Maps © Google; Map of Antarctica: Landsat LIMA Mosaic © USGS, NASA, BAS, NFS.}
	\label{fig:front_overview}
\end{figure*}
\section{Evaluation}\label{sec:evaluation}
\subsection{Dataset}\label{sec:dataset}
\subsubsection{Study Sites}
We selected outlet glaciers at the Antarctic Peninsula (AP) 
to train and evaluate the glacier front mapping on SAR intensity imagery using CNN.
The AP is a hot spot of global climate change. A significant temperature increase was reported for the \nth{20} century~\cite{oliva2017recent, turner2016absence} 
strongly affecting its ice masses. Pronounced retreat and even disintegration of some ice shelves along the AP was observed~\cite{cook2010overview}. 
Most of the northern ice shelves at the Prince-Gustav-Channel (PGC) and Larsen-A embayment disintegrated in 1995, and at the Larsen-B embayment in 2002~\cite{cooper1997historical, scambos2004glacier, skvarca1998evidence}. 
The reduction of the buttressing forces led to a strong increase in ice discharge and further frontal retreat of the former ice shelf tributary glaciers~\cite{rott2014mass, seehaus2015changes}. 
Due to its relevance and the data availability, we selected several of these former ice shelf tributaries for our analysis (see \cref{fig:front_overview}).

\begin{table*}[hbt]
	\centering
		\caption{Overview of the SAR sensors and specifications used to acquire the SAR images of the Antarctic Peninsula dataset.}
	\label{tab:sar_sensors}
	\begin{tabular}{cccccccc}
		\toprule
		Platform	& Sensor	& Mode	& SAR band	& \makecell{Repetition\\cycle\\{}[d]}	& Time interval	& \makecell{Multi\\looking\\factor}	& \makecell{Ground range\\resolution\\{}[m]} 	\\ \midrule
		ERS-1/2	& SAR	& IM	& C band	& 35/1	& \makecell{13. November 1995 -\\ 26. February 2010}	& 1x5	& 20	\\
		\makecell{RADARSAT 1 \\ (SLC \& PRI\\format)}	& SAR	& ST	& C band	& 24	& \makecell{10. September 2000 -\\ 20. January 2008}	& 1x4	& \makecell{20 (SLC) \\ 12.5 (PRI)}	\\
		Envisat	& ASAR	& IM	& C band	& 35	& \makecell{05. December 2003 -\\ 03. July 2010}	& 1x5	& 20	\\[0.3cm]
		ALOS	& PALSAR	& FBS	& L band	& 46	& \makecell{18. May 2006 -\\ 03. March 2011}	& 2x5	& 16,7	\\
		\makecell{TerraSAR-X \\ TanDEM-X \\ (SLC format)}	& SAR	& SM	& X band	& 11	& \makecell{13. October 2008 -\\ 22. December 2014}	& 3x3	& 6.7	\\ \bottomrule
	\end{tabular}
\end{table*}
The glacier system of the Sjögren-Inlet (SI) and Dinsmoore-Bombardier-Edgworth (DBE) glacier systems were major tributaries to the PGC and Larsen-A ice shelves, respectively. \etal{Seehaus}~\cite{seehaus2015changes, seehaus2016dynamic} 
carried out a detailed analysis of the glacier evolution after the disintegration of the ice shelves. 
Simultaneously, the glacier fronts strongly retreated after the ice shelf disintegration and stabilized at SI glacier system after about 15 years, whereas at DBE, the glacier re-advance and short-term recession was observed.
Crane, Mapple and Jorum glaciers are former tributaries of the Larsen-B ice shelf. \etal{Rott} reported a similar reaction to the ice shelf break-up as DBE and SI glacier systems~\cite{rott2014mass, tc-12-1273-2018}. 

We used imagery from the satellite missions ERS-1/2, Envisat, RadarSAT-1, ALOS, TerraSAR-X (TSX) and TanDEM-X (TDX), covering the period 1995--2014. The German Aerospace Center (DLR), the European Space Agency (ESA) and the Alaska Satellite Facility (ASF) provided the SAR data for our analysis, via granted data proposals. The SAR acquisitions were multilooked to reduce speckle noise and calibrated. The ASTER digital elevation model of the AP~\cite{cook2012new} was used for the geocoding and the orthorectification of the images. The specifications and parameters of the SAR sensors and imagery are provided in \cref{tab:sar_sensors}. The GAMMA RS Software was used for processing the imagery.

%

\subsubsection{Data Generation}
\begin{figure}
	\centering
	\begin{subfigure}[b]{0.32\linewidth}
		\centering
		\includegraphics[width=1\linewidth]{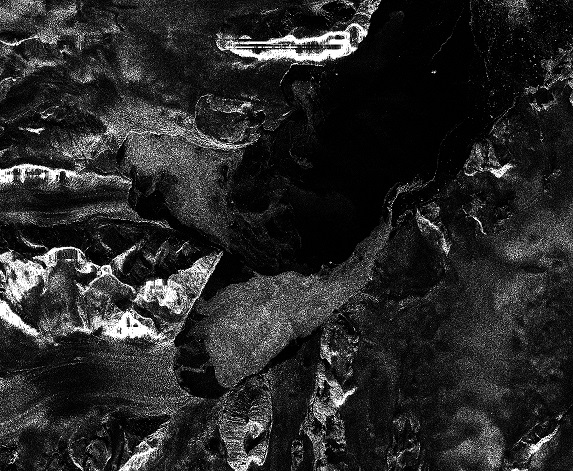}
		\caption{}\label{fig:gt_a}
	\end{subfigure}
	\begin{subfigure}[b]{0.32\linewidth}
		\centering
		\includegraphics[width=1\linewidth]{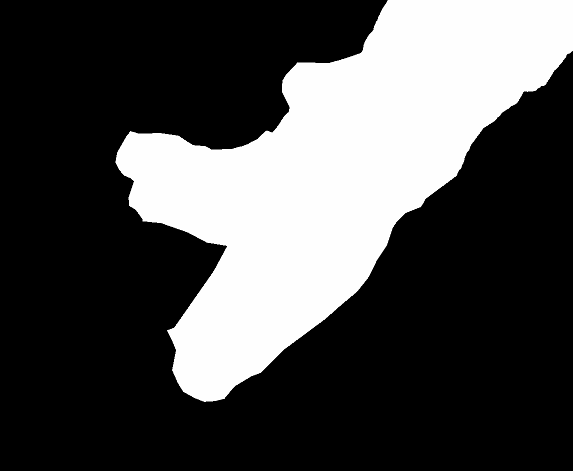}
		\caption{}\label{fig:gt_b}
	\end{subfigure}
	\begin{subfigure}[b]{0.32\linewidth}
		\centering
		\includegraphics[width=1\linewidth]{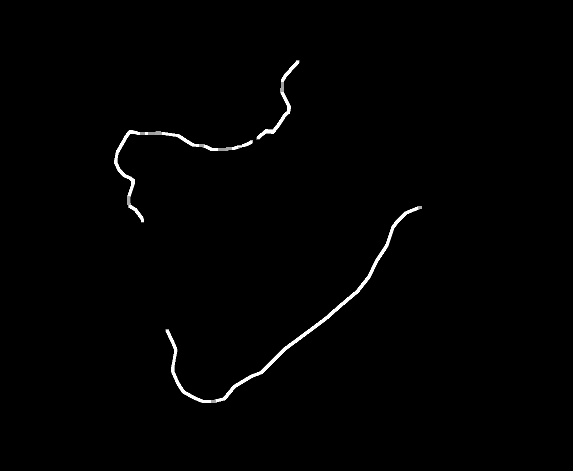}
		\caption{}\label{fig:gt_c}
	\end{subfigure}
	\caption{Statistical Thresholding post-processing steps: \subref{fig:gt_a} predicted distance map, \subref{fig:gt_b} binarized distance map, and \subref{fig:gt_c} calving front line.} 
	\label{fig:gt}
\end{figure}

The manually annotated calving front positions for the DBE and SI glacier systems have been generated and used by \etal{Seehaus}~\cite{seehaus2015changes, seehaus2016dynamic}. Additionally, we manually annotated the glacier front positions of the Larsen-B embayment. \Cref{fig:gt} depicts the ice mélange and glaciers at the non-ice mélange regions (referred to as \emph{zone} in this work), and the glacier calving front (referred to as \emph{line} in this work) ground truth images for a sample SAR image in our dataset.

A quality factor, ranging from $1$ to $6$, was assigned to each SAR image. The quality factors are a subjective measure of the reliability of the picked front position depending on the similarity of the ice mélange and the glacier, with $1$ being the most reliable and $6$ the most unreliable. \Cref{tab:quality_factors} shows the quality factors and the respective uncertainty values, perpendicular to the glacier front. 
In this study, we discarded the glacier fronts with a quality factor of $6$.
\begin{table}[tbp]
	\centering
	\def\arraystretch{1.5}
	\begin{tabular}{ c c c  c c c c}
		\toprule
		\makecell{Quality  \\ factor}& 1	& 2	& 3	& 4	& 5	& 6	 \\
		\midrule
		
		$\sigma_{f}(m)$ & 70 & 130 & 200 & 230 & 450 & -- \\
		\bottomrule
	\end{tabular}
	\caption{Quality factors of manual glacier front detection and the assigned horizontal accuracies $\sigma_{f}$ (perpendicular to estimated glacier front line).}
	\label{tab:quality_factors}
\end{table}

Subsets of the preprocessed SAR intensity imagery were generated covering the areas of interest (see \cref{fig:front_overview}) and converted to 16-bit single-channel images. Each manually picked glacier front position (spatial line) was combined with the coastline and rock-outcrop polygons (based on the Antarctic Digital Database at the AP) to generate a categorized 8bit single-channel image.
The ground truth images have the same size and resolution as the respective SAR image subsets and are used for training the models.

\subsection{Experiments}
First, we compute the distance maps from the glacier front lines (ground truth). We set the decay parameter $\gamma=7$, which produced the best result within the set $\gamma \in \{1\dots10\}$. The same decay value is also used by \etal{Meyer}~\cite{meyer2018pixel}. The SAR images and the distance maps are then resized to $512\times512$ using linear interpolation and used as input for the first U-Net.
The contraction path consists of five down-sampling steps with two $5\times5$ convolutional layer, each followed by a batch normalization layer and leaky rectified linear unit (Leaky ReLU) activation function, as well as a $2\times2$ max pooling operation. Each block of the expansion path uses a $4\times4$ transposed convolution layer with a stride of $2\times2$ that gets concatenated with the corresponding feature map of the contracting path. Then the concatenation result goes through the same combination of two $5\times5$ convolutional layers, batch normalization layers, and Leaky ReLUs analogous to the contraction path. The number of the convolutional filters gets doubled with each down-sampling step from $32$ to $1024$, and correspondingly halved in each up-sampling step from $1024$ to $32$. The final layer is a $3\times3$ convolution layer with a Sigmoid activation function. The network's weights are initialized using He-initialization~\cite{He_2015_ICCV}.

We train the U-Net from scratch with $310$ training images with a batch size of $3$. Because of the limited training set size, we augment the images by horizontal flips and rotations of $90$, $180$, and $270$ degrees, which enlarges our training set by the factor of eight. The training is terminated when the loss value of the validation set, which includes $78$ images, does not decrease for $30$ consecutive epochs. We use Adam optimizer with a learning rate of $0.001$ to minimize the mean-squared error (MSE) loss function. The training lasted for $83$ epochs and resulted in a validation MSE of $0.003 \pm 0.001$. The model shows a visually consistent and accurate distance map prediction results across the validation and test data set. 

In the next step, we resized the predicted distance map back to the original image size. We use bicubic interpolation to implement a smooth transformation with less distortion. We apply the different post-processing approaches, \cf \cref{sec:post-processing}, to extract the final calving front line. The three different post-processing algorithms are parameterized as follows:
\subsubsection{Statistical Thresholding} 
By setting the percentage of pixels for the threshold value for each image to \SI{95}{\percent}, most of the lower values from the distance map get removed. This results in a narrow shape mask containing our predicted calving front.
	
\subsubsection{Conditional Random Field} 
As for the majority of the parameter values in the CRF model, we followed \etal{Chen}~\cite{DBLP:journals/corr/ChenPK0Y16}. However, we increase the standard deviation of the Bilateral kernel to $\theta_{\alpha}=512$, which overall decreases the number of artifacts and narrows the spread of the generated glacier front. We also removed the influence of the Gaussian potential, as it further reduced the width of the generated fronts and achieved the most accurate front line.

\subsubsection{Second U-Net}
The model is trained with distance maps predicted by the pixel-wise regression model with the objective of segmenting the calving fronts. We bypass the class-imbalance problem by morphologically dilating the ground truth calving front line with a $5\times5$ kernel as the structuring element (SE). The same dataset split is preserved to ensure a comparable evaluation. In the optimization process, we used the Adam optimizer with the learning rate of $\text{lr}=5e-6$ to minimize the binary cross-entropy (BCE) loss function. 
We use the same U-Net architecture and initialize it analogously to the previous model. Using the same early stopping policy as before, the training took $16$ epochs. 

The individual approaches above show different results. n particular, the width of the predicted glacier front lines varies significantly from one approach to the other.
To ensure a fair comparison, we adjusted the width of the predictions. 
Since both the CRF and the Second U-Net methods predict thicker glacier fronts, we reduced them for the evaluation phase. The output of the Statistical Thresholding method has a width of a single pixel (skeleton), hence we expand it accordingly. We use morphological dilation and erosion operators to adjust the width of the predicted glacier calving front lines. In the width adjustment process, we searched for the optimum structuring element (SE) size resulting in a small difference in the sum of calving front pixels for different approaches on a few test images and then applied it to the entire evaluation. 

We compare our results with the state-of-the-art calving front line segmentation approach proposed by \etal{Zhang}~\cite{zhang2019automatically}, where each image is subdivided into $512\times512$ patches. Those are individually segmented by the U-Net. Afterwards, the output patches are combined to fit the original image size, which results in the final segmentation output. Their proposed U-Net is trained to segment the whole ice mélange and non-ice mélange regions and the calving front line is extracted via post-processing. In our application of this method, the U-Net prediction is binarized to extract the glacier and ice mélange regions. The prediction includes many artifacts. Therefore, the small holes and gaps are filled within the predicted segmentation mask using morphological closing and the remaining artifacts are removed by extracting the largest connected component (CC) and discarding the rest. The edge of this CC defines the glacier calving front line, which was automatically extracted using Canny edge detection algorithm~\cite{Canny}.
The U-Net of \etal{Zhang}~\cite{zhang2019automatically} is trained with the extracted patches using Adam optimizer with $\text{lr}=1e-5$ and the BCE loss function. Similar to our training strategy, the training was monitored by an early stopping mechanism with $30$ epochs patience and converged after $103$ epochs. We refer to this method as \emph{Baseline zone}. 
Moreover, we also trained their U-Net architecture using only the glacier calving front line as the ground truth. We thickened the glacier calving front lines in the $512\times512$ ground truth image patches using a morphological dilation with a structuring element of size $5\times5$. The training converged after $78$ epochs. This approach is referred to as \emph{Baseline line}.

\begin{table}
	\centering
	\begin{tabular}{@{}ccccccc@{}}
		\toprule
		\multirow{3}{*}{Score}		& \multirow{3}{*}{\makecell{Tolerance \\ in meters}}	& \multicolumn{2}{c}{Zhang et al.~\cite{zhang2019automatically}}	& \multicolumn{3}{c}{Proposed Methods}	\\ \cmidrule{3-4}\cmidrule{5-7}
		&& \makecell{Baseline\\zone} & \makecell{Baseline\\line} & Threshold	& CRF	& \makecell {Second\\U-Net}	\\
		\midrule
		\multirow{6}{*}{Dice} 		
		& 0 	& 0.0177	& 0.0768	& 0.1176 	& 0.1439	& \bf 0.1603	\\
		& 50 	& 0.0756	& 0.2004	& 0.2656 	& 0.3065	& \bf 0.3245 	\\
		& 100 	& 0.1233	& 0.2678	& 0.4043 	& 0.4438	& \bf 0.4629 	\\
		& 150 	& 0.1615	& 0.3167	& 0.5233 	& 0.5544	& \bf 0.5702 	\\
		& 200 	& 0.1838	& 0.3405	& 0.5908 	& 0.6145	& \bf 0.6296 	\\
		& 250 	& 0.2072	& 0.3628	& 0.6529 	& 0.6695	& \bf 0.6835 	\\
		
		\midrule
		
		\multirow{6}{*}{IoU} 		
		& 0 	& 0.0089	& 0.0410	& 0.0642	& 0.0795	& \bf 0.0908 	\\
		& 50 	& 0.0407	& 0.1192	& 0.1555	& 0.1826	& \bf 0.1958	\\
		& 100 	& 0.0699	& 0.1702	& 0.2592	& 0.2891	& \bf 0.3063 	\\
		& 150 	& 0.0945	& 0.2099	& 0.3625	& 0.3893	& \bf 0.4055 	\\
		& 200 	& 0.1097	& 0.2306	& 0.4285 	& 0.4500	& \bf 0.4668 	\\
		& 250 	& 0.1256	& 0.2499	& 0.4932 	& 0.5096	& \bf 0.5260 	\\
		\bottomrule
	\end{tabular}
	\caption{Quantitative results of the proposed methods alongside the state-of-the-art methods~\cite{zhang2019automatically}.}
	\label{tab:quan_ev}
\end{table}
\begin{figure*}
	\centering
	\begin{subfigure}[b]{0.19\linewidth}
		\centering
		\includegraphics[width=1\linewidth]{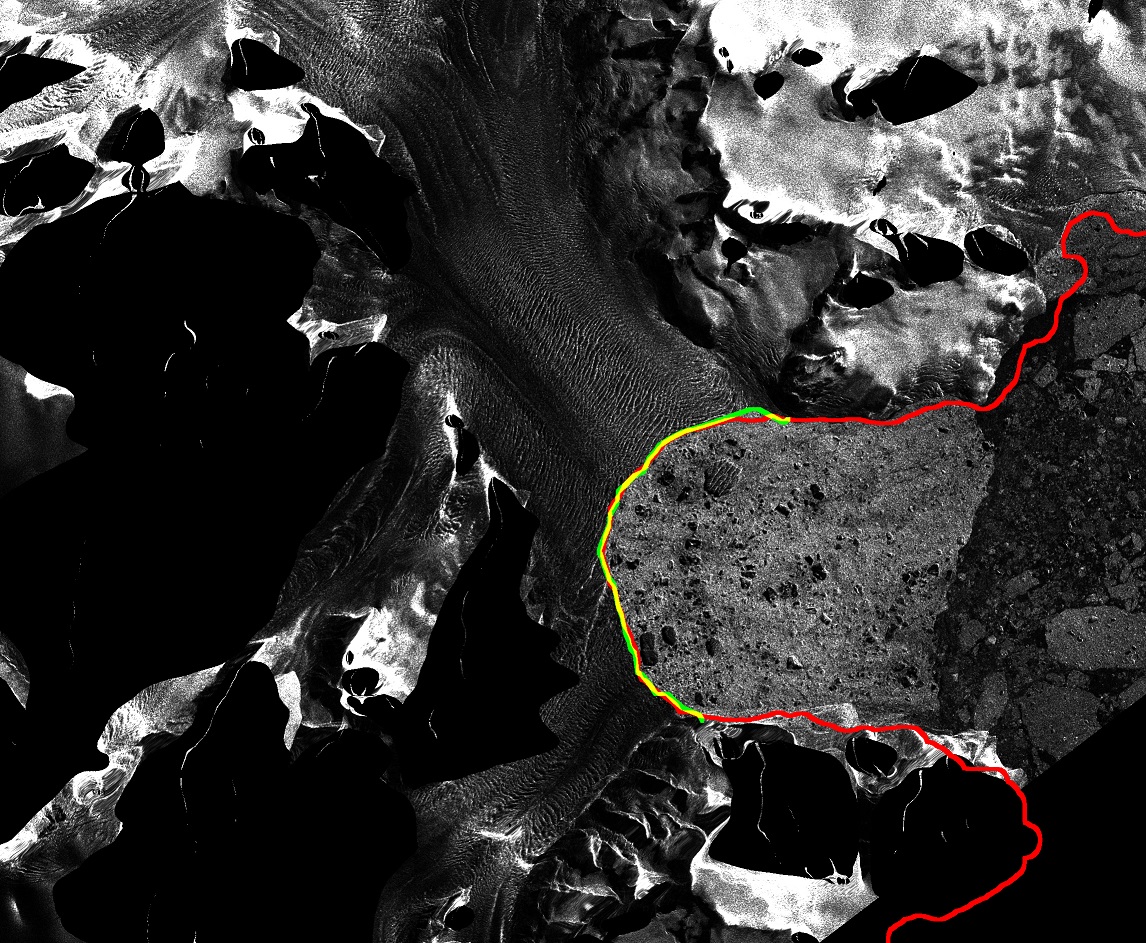}
		\vspace{1pt}
		
		\includegraphics[width=1\linewidth]{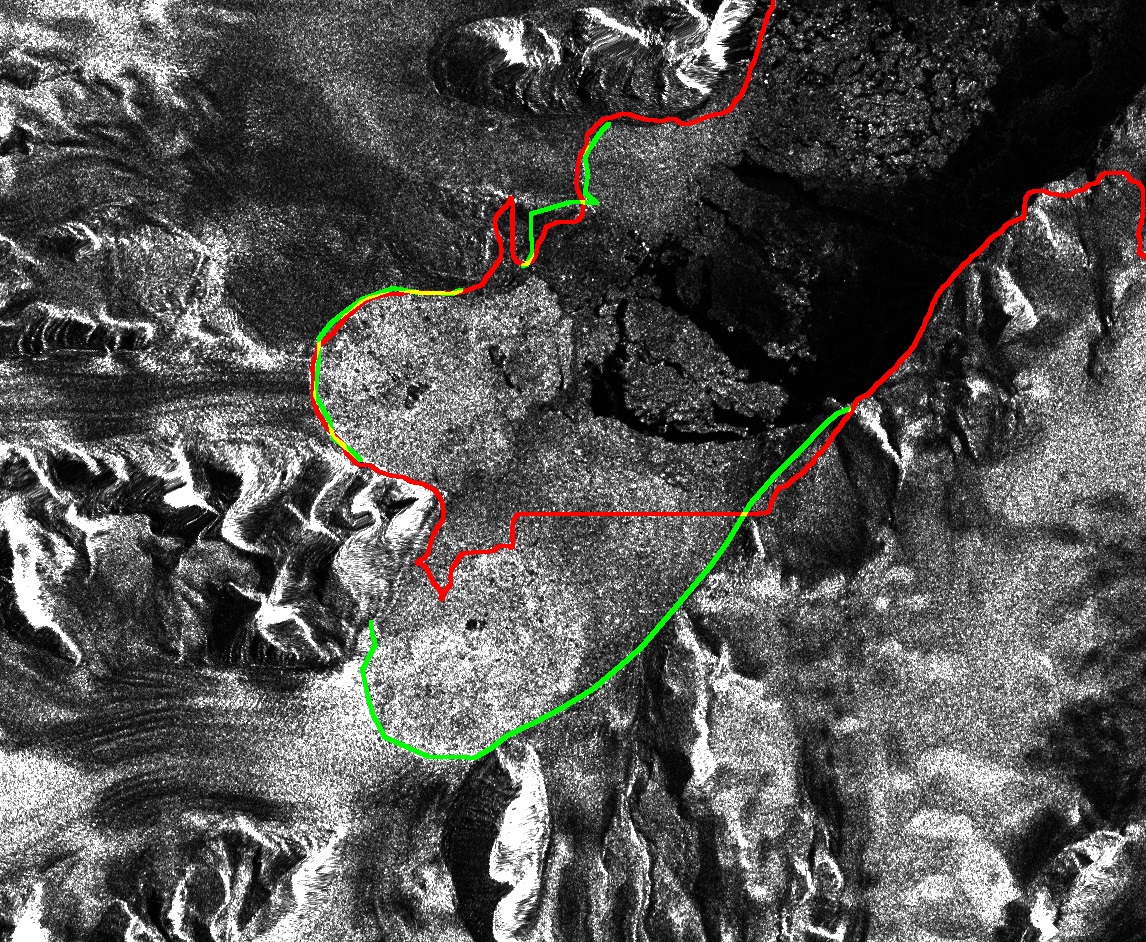}
		\caption{Baseline zone}\label{fig:qual_ev_a}
	\end{subfigure}
	\medskip
	\begin{subfigure}[b]{0.19\linewidth}
		\centering
		\includegraphics[width=1\linewidth]{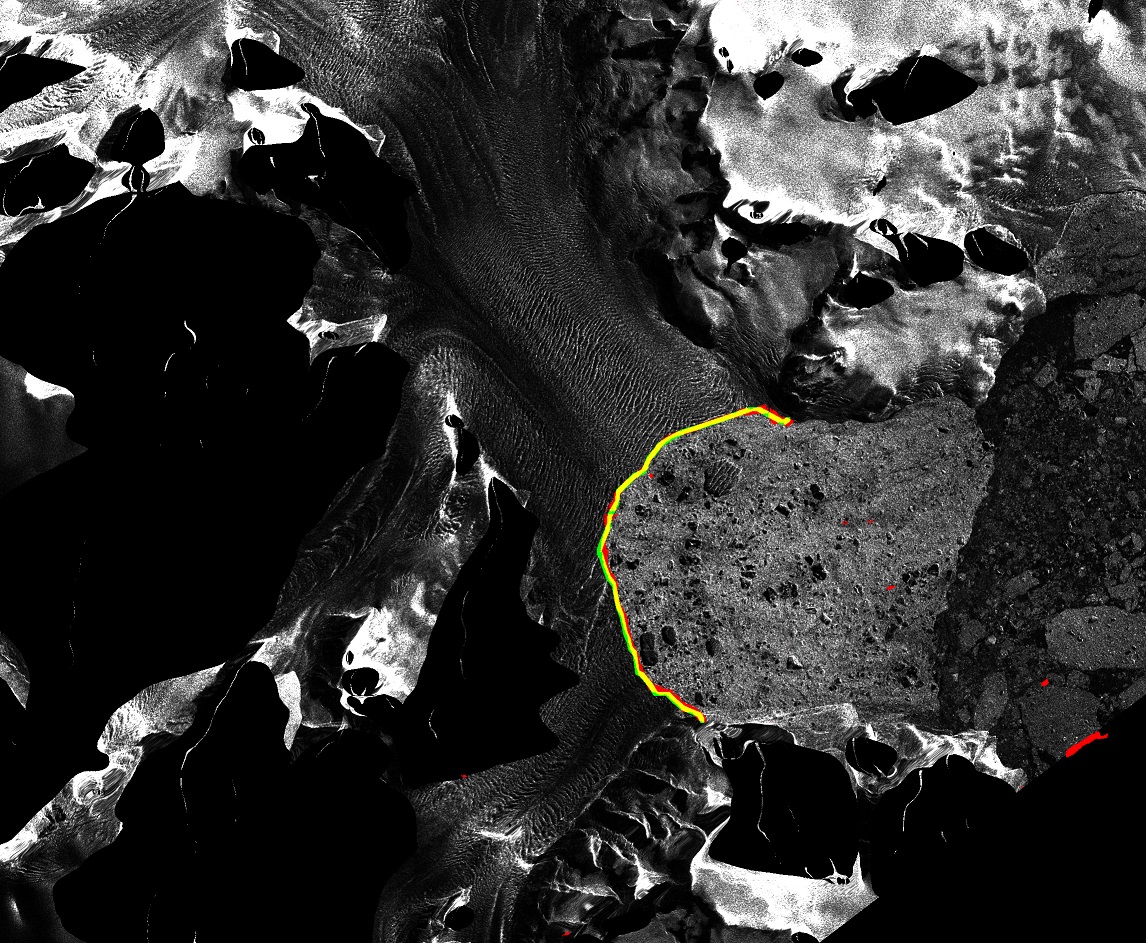}
		\vspace{1pt}
	
		\includegraphics[width=1\linewidth]{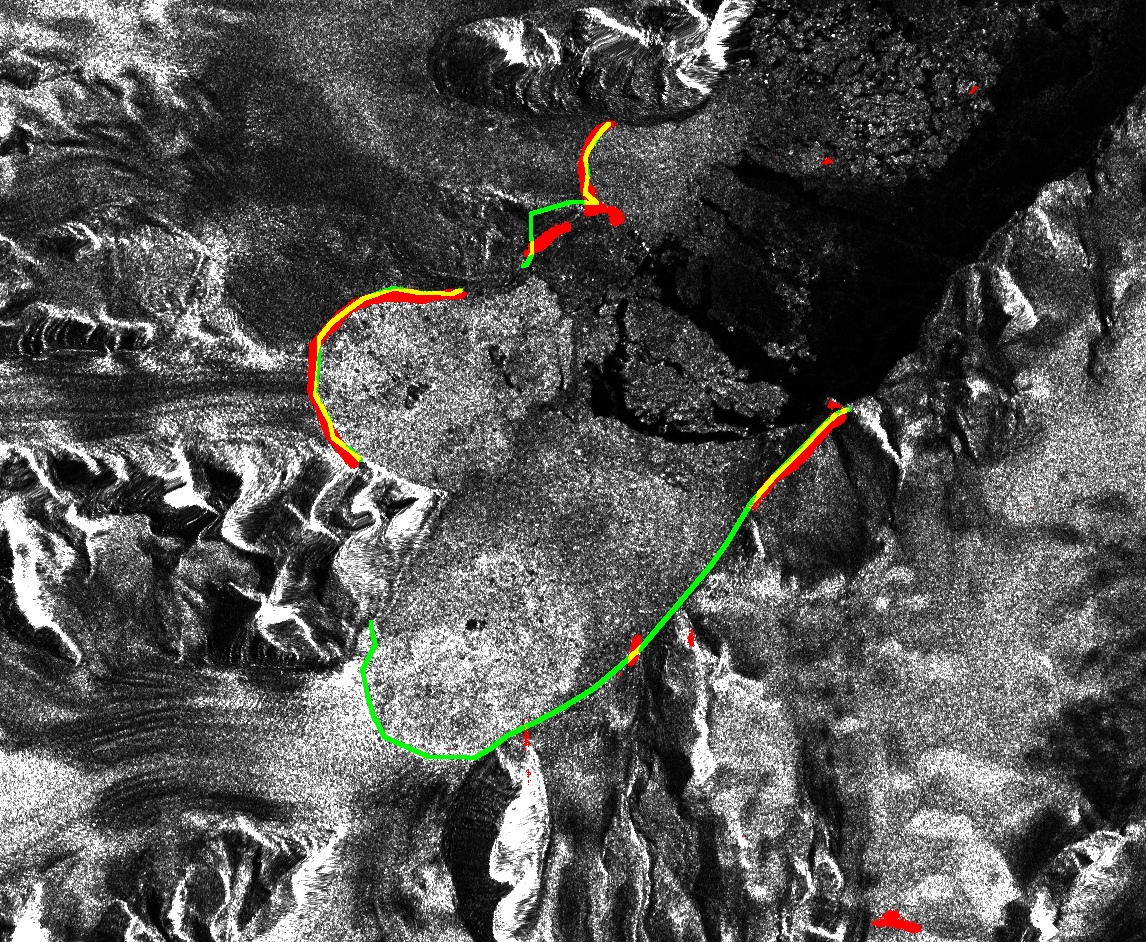}
		\caption{Baseline line}\label{fig:qual_ev_b}
	\end{subfigure}
	\medskip
	\begin{subfigure}[b]{0.19\linewidth}
		\centering
		\includegraphics[width=1\linewidth]{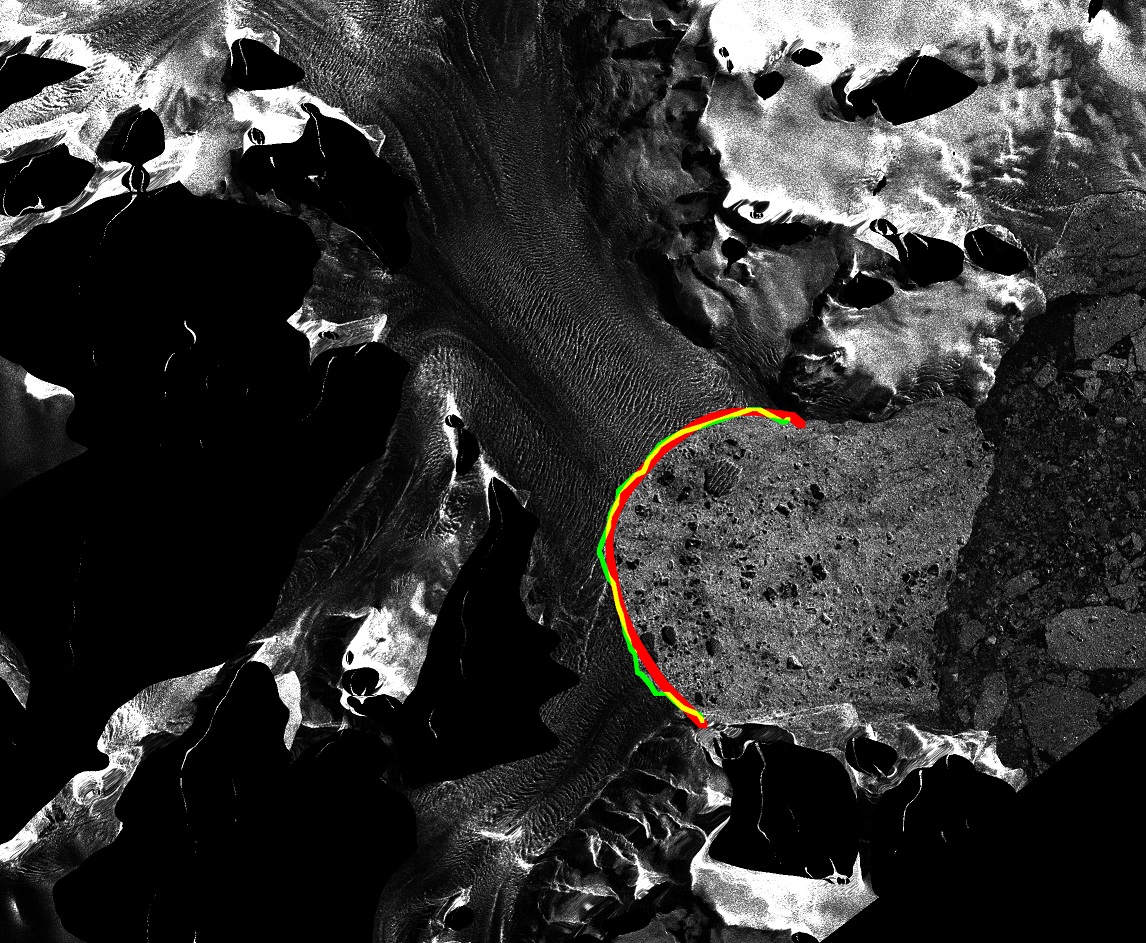}
		\vspace{1pt}
		
			\includegraphics[width=1\linewidth]{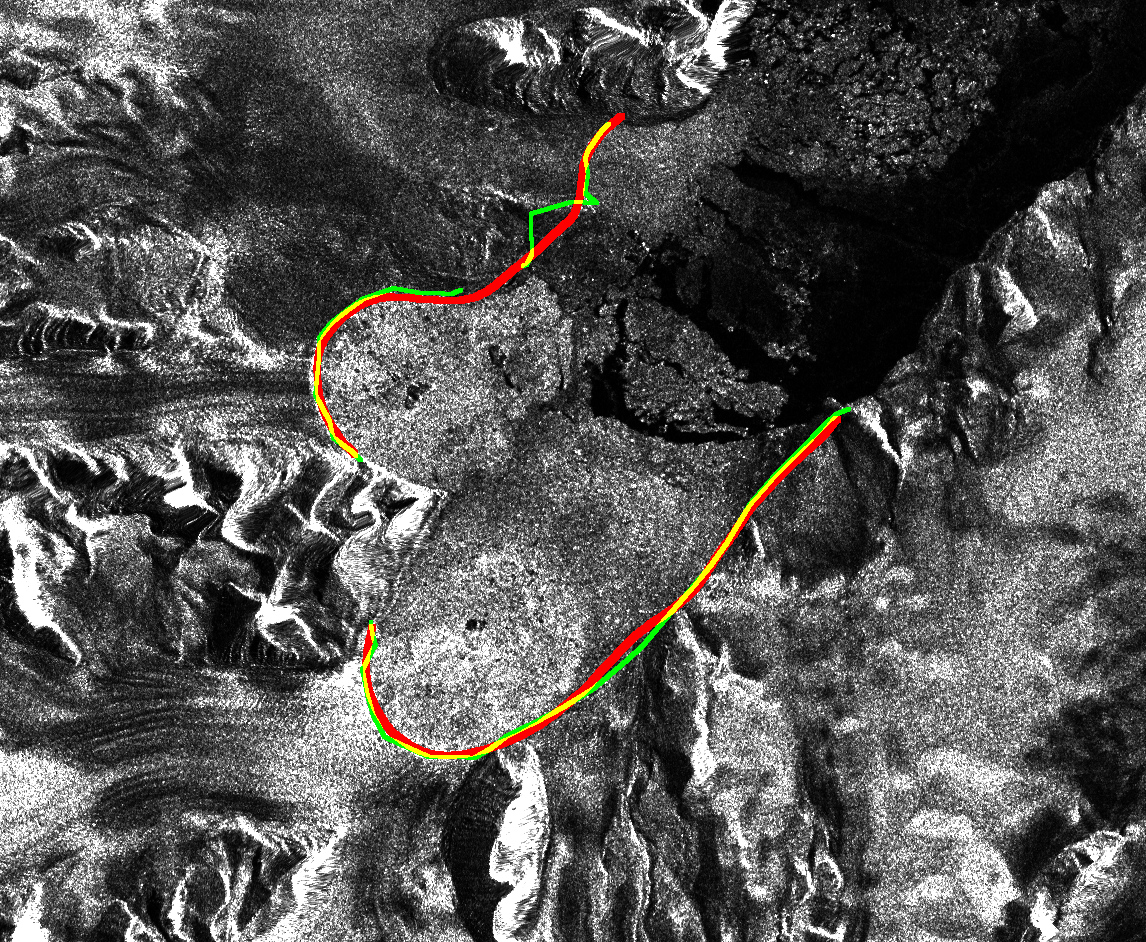}
			\caption{Statistical Thresholding}\label{fig:qual_ev_c}
	\end{subfigure}
	\medskip
	\begin{subfigure}[b]{0.19\linewidth}
		\centering
		\includegraphics[width=1\linewidth]{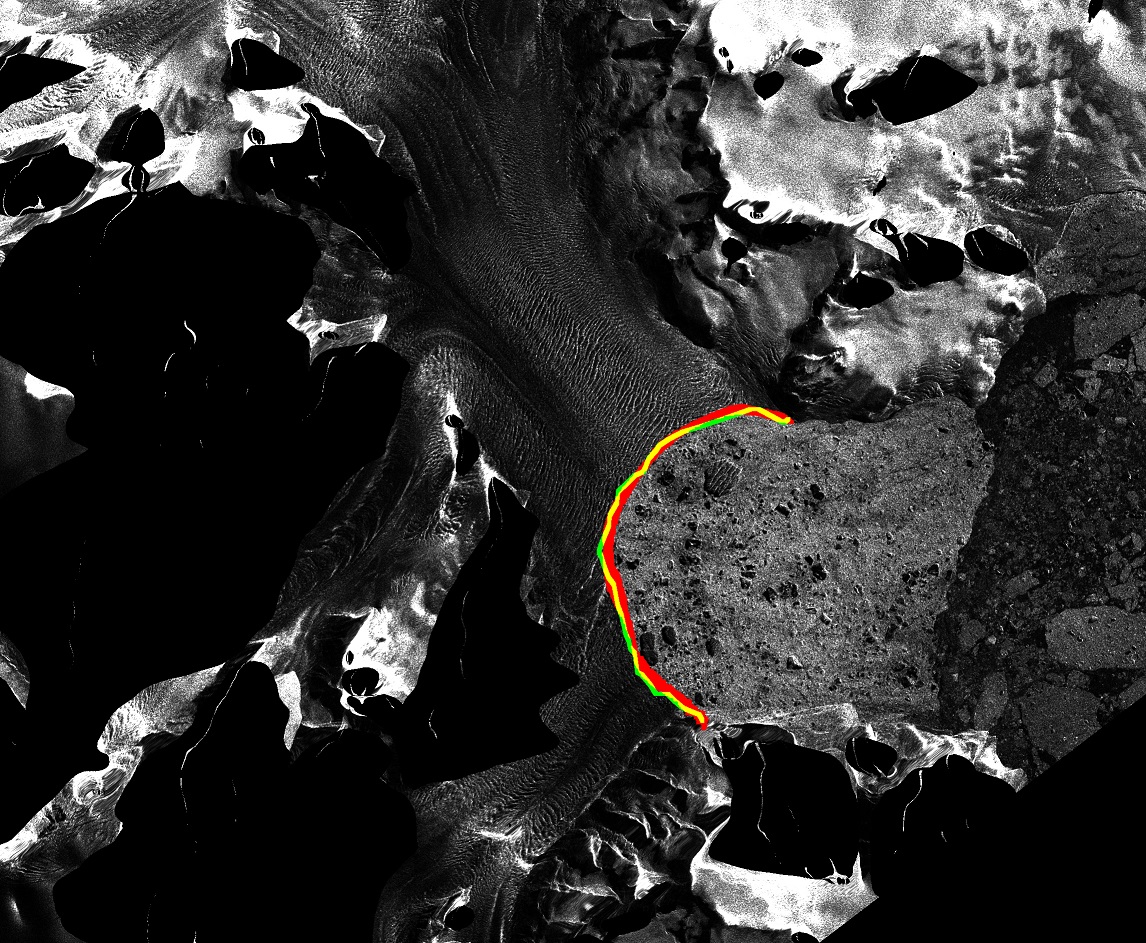}
		\vspace{1pt}
		
			\includegraphics[width=1\linewidth]{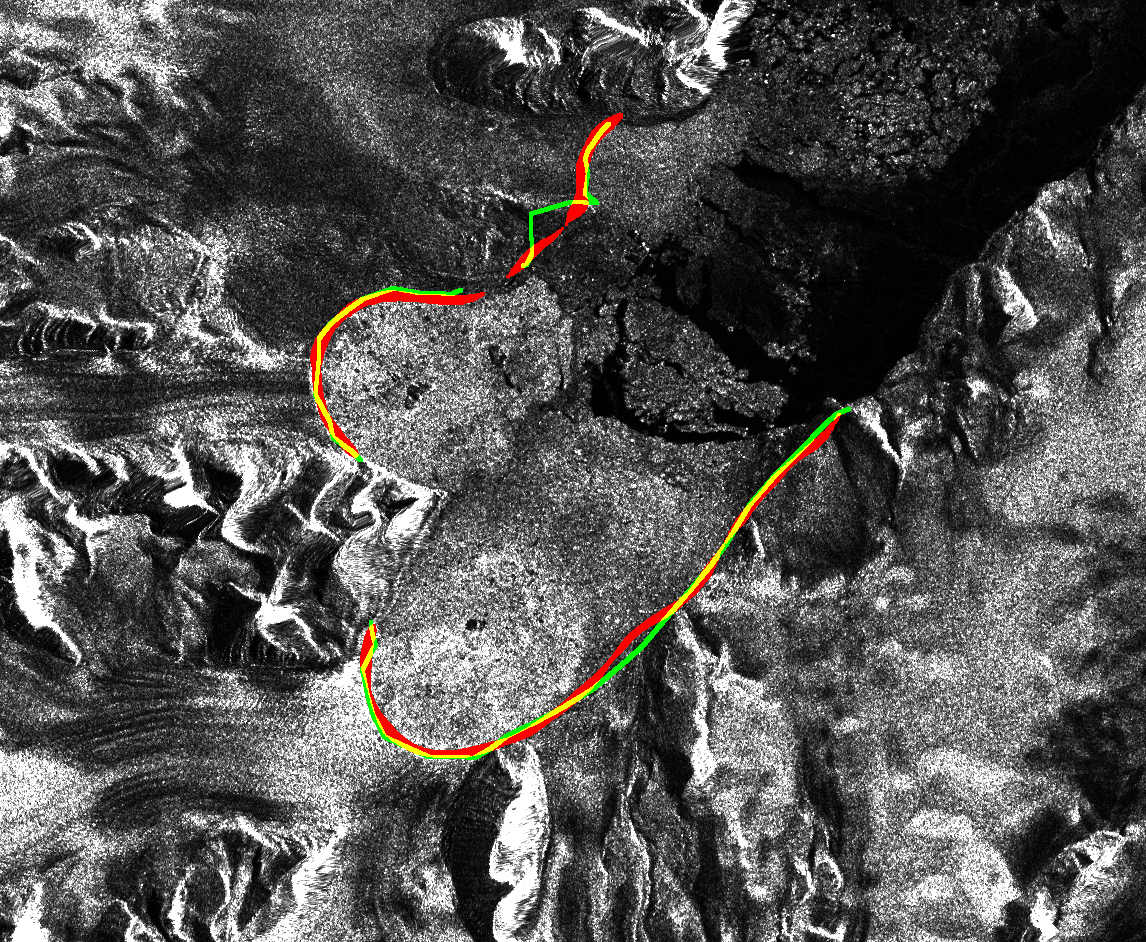}
			\caption{CRF}\label{fig:qual_ev_d}
	\end{subfigure}
	\medskip
	\begin{subfigure}[b]{0.19\linewidth}
		\centering
		\includegraphics[width=1\linewidth]{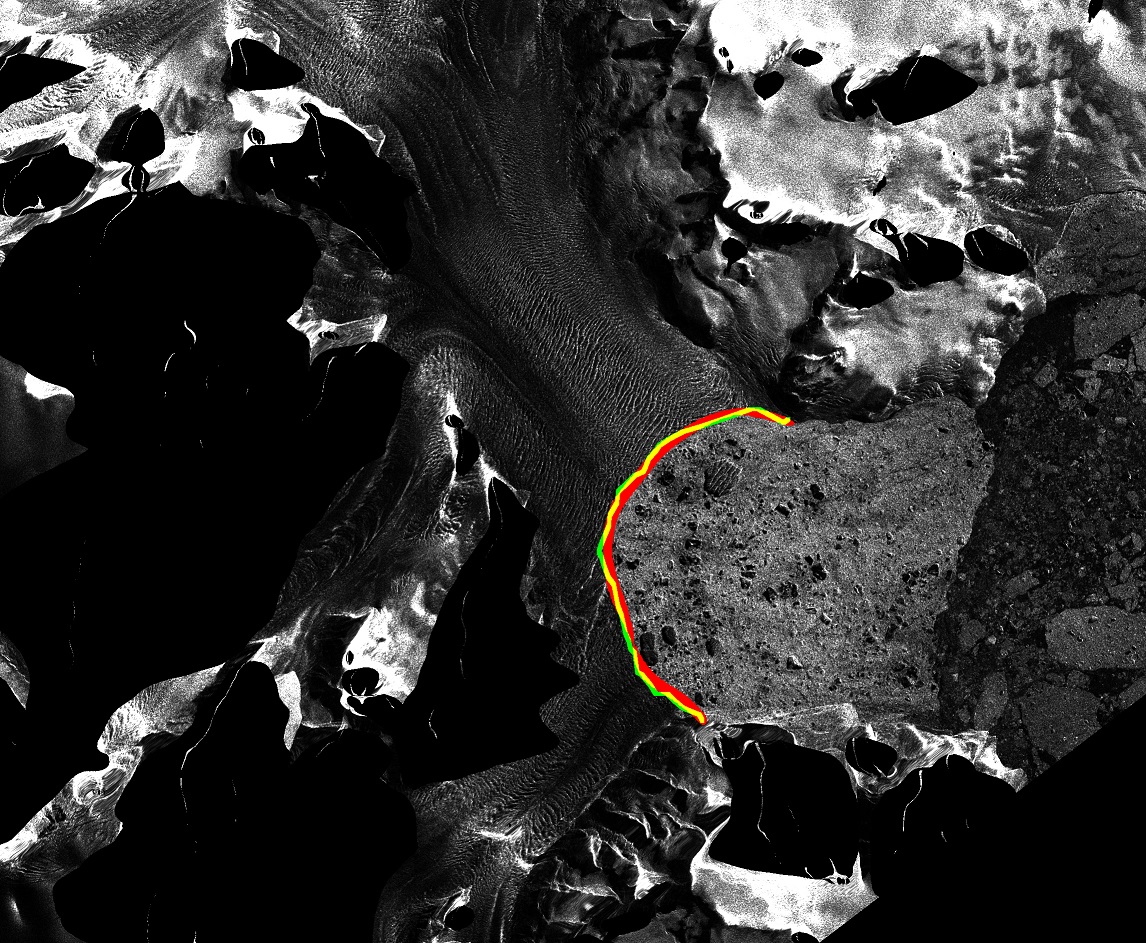}
		\vspace{1pt}
		
		 \includegraphics[width=1\linewidth]{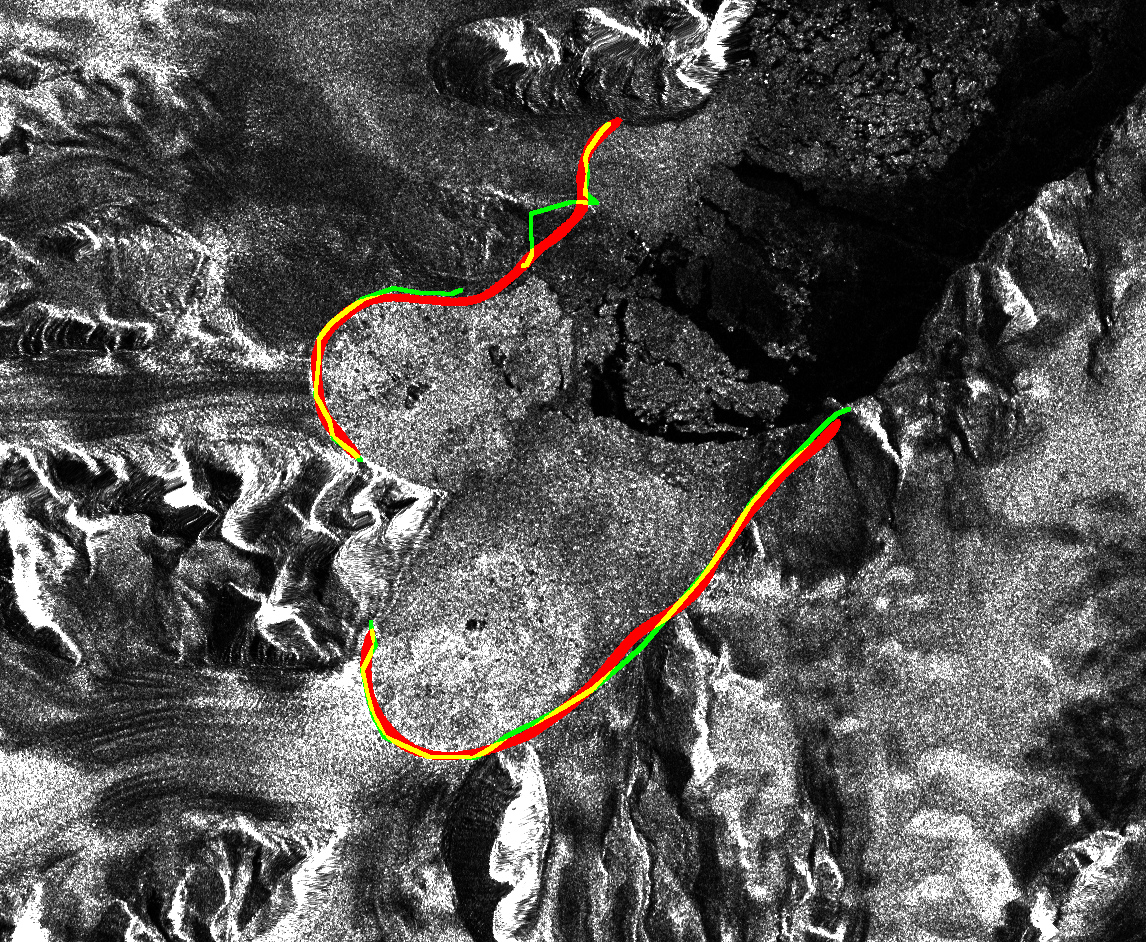}
		 \caption{Second U-Net}\label{fig:qual_ev_e}
	\end{subfigure}
	
	\caption{Qualitative results of the different approaches for two sample glaciers. Green color represents the ground truth label, red shows the prediction, and yellow illustrates the correctly predicted pixels, \ie the overlap between the prediction and the ground truth.} 
	\label{fig:qual_ev}
\end{figure*}

\paragraph{Quantitative Evaluation}
We report the dice coefficient and intersection over union (IoU) metrics. Since we try to predict a segmentation mask with a single pixel width, a small deviation in the prediction line severely affects the results leading to numerically low initial scores. We, therefore, allow marginal deviation from the ground truth calving front line by introducing a tolerance factor. This factor defines a distance between the predicted line and the actual glacier front that will not be penalized. The tolerance is given in meters, which is translated into pixels using the individual SAR images' spatial resolution. We implement the tolerance using morphological dilation $\delta$ with a structuring element $B$. The structuring element's size $s$ is determined from the tolerance value and the spatial resolution of the image. We dilate the predicted and the ground truth calving front and then calculate dice coefficient with different tolerances:
\begin{equation}
	\text{dice}_{s}(y_{\text{true}}, y_{\text{pred}}) = \frac{\delta_B(y_{\text{true}}) * \delta_B(y_{\text{pred}}) }{\delta_B(y_{\text{true}}) + \delta_B(y_{\text{pred}})} \;.
\end{equation}
The IoU is related to the Dice coefficient by $\text{IoU}={\text{dice}}/{(2-\text{dice})}$ and is just given for reference.

The quantitative results can be observed in \cref{tab:quan_ev}. Since the manually annotated glacier calving front lines in our dataset with quality factor $1$ to $4$ have up to $230$ meters accuracy uncertainty, we were tolerant to up to 250 meters of deviation from the ground truth line and therefore, carried out our evaluation for the wide tolerance range of $0$, $50$, $100$, $150$, $200$, and $250$ meters. As expected, higher tolerances consistently result in higher performance metric values. In almost all scenarios, increasing the tolerance from $0$ meters to $50$ meters increases the dice coefficient (and IoU) between \SI{100}{\percent} and \SI{200}{\percent}.
As far as the state-of-the-art methods are concerned, training the U-Net to learn the glacier front lines directly (\emph{Baseline line}) resulted in significantly better performance than \emph{Baseline zone}, \ie segmenting the ice-mélange and non-ice mélange, and extracting the boundary between them.
In comparison to the state-of-the-art methods, all the proposed methods in this work outperform both \emph{Baseline zone} and \emph{Baseline line}. For example for the tolerance of $150$ meters, the Statistical Thresholding and the Second U-Net methods achieved $0.52$ and $0.57$ dice coefficient while \emph{Baseline line} obtains $0.32$.
Among the proposed methods, CRF and Second U-Net perform on average \SI{7}{\percent} to \SI{11}{\percent} better than the Statistical Thresholding approach. The Second U-Net shows the best quantitative results, but note that training a second network is computationally more expensive compared to the hyperparameter selection for the Statistical Thresholding and CRF approaches. 

\paragraph{Qualitative Evaluation}
\Cref{fig:qual_ev} depicts the automatically delineated glacier calving front positions using the proposed methods for two sample images in the test set. Each image illustrates the predicted glacier front in red, the ground truth in green, and their overlap (correct prediction) in yellow, overlaid on their corresponding SAR image.
Analogous to the quantitative results, all the proposed pixel-wise regression-based models consistently predict the calving front position more accurately compared to the state-of-the-art methods. Our methods correctly predict most of the false alarms of the \emph{Baseline zone} and \emph{Baseline line} approaches and do not suffer from severe pollution due to outliers in the predictions. 
Among our proposed methods, the shape and structure of the glacier calving fronts generated by the Thresholding method are often simpler than the other two delineation approaches. The Second U-Net post-processor shows the best qualitative results since its prediction closely matches the manually annotated calving front, which is backed up consistently with our quantitative evaluation results. 

\Cref{zoom1-zoom} highlights the mispredictions by the \emph{Baseline line} method~\cite{zhang2019automatically}, which are corrected by the Second U-Net approach proposed in this work, see \cref{zoom2-zoom}.
The state-of-the-art approach output suffers from visible artifacts, which are far off from the ground truth and do not represent the actual calving front. Further, the many green areas show the inconsistency within the prediction. Our proposed methodology using the Second U-Net tracks the ground truth consistently and avoids false alarms.
%
\begin{figure*}
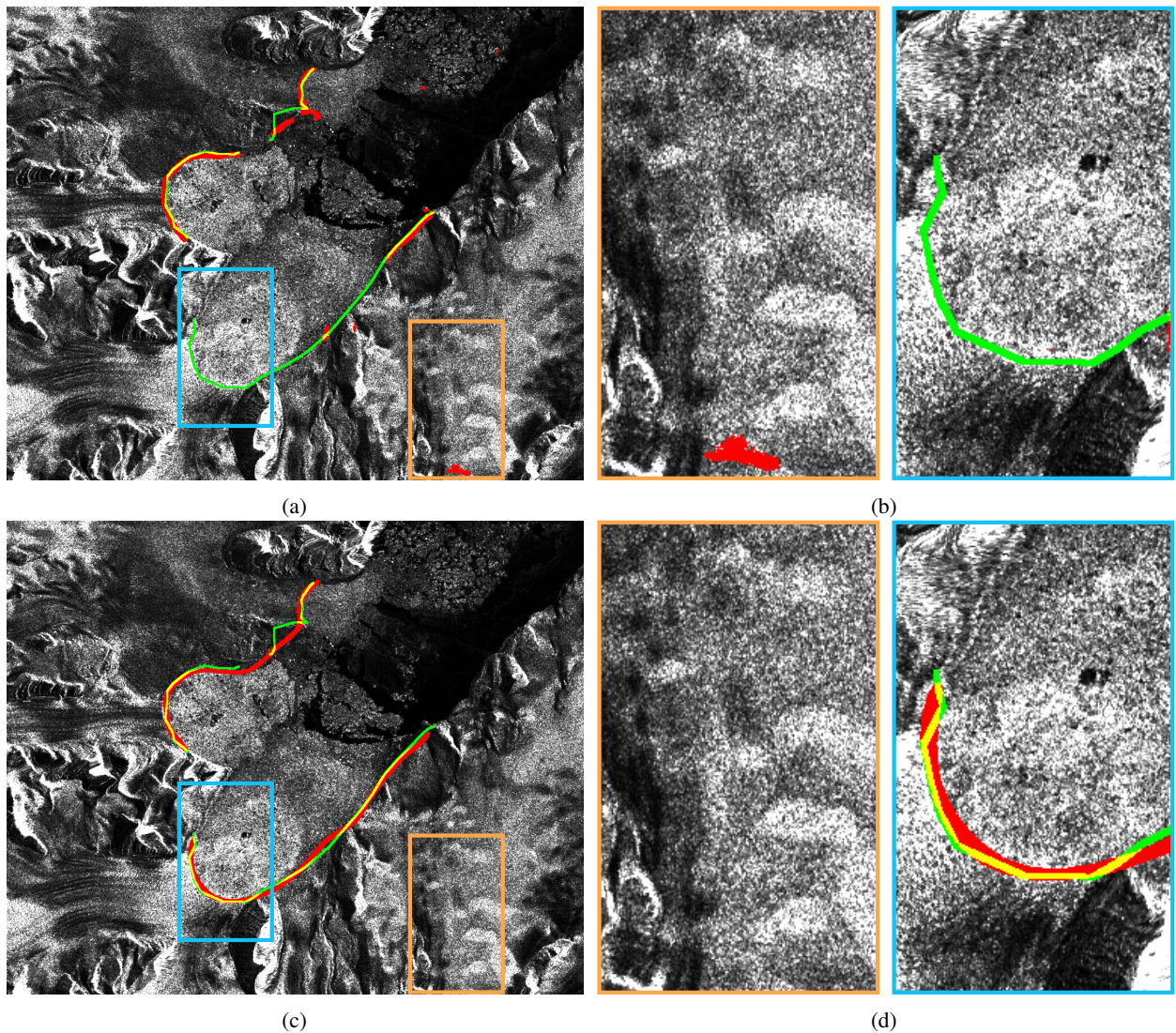

	\centering
	\begin{tikzpicture}[figurename=zoom1,zoomboxarray, 
		zoomboxarray columns=2, 
		zoomboxarray rows=1, 
		zoombox paths/.append style={ultra thick}
		]
		\node [image node] { \includegraphics[width=0.45\linewidth]{Figures/final/patch_front_dilate/2006-05-06_RSAT_20_3.jpg}};
		\zoombox[color code=orange!70, magnification=3]{0.78,0.17}
		\zoombox[color code=cyan!70, magnification=3]{0.38,0.28}
	\end{tikzpicture}

	\begin{tikzpicture}[figurename=zoom2,zoomboxarray, 
		zoomboxarray columns=2, 
		zoomboxarray rows=1, 
		zoombox paths/.append style={ultra thick, orange!75}
		]
		\node [image node] { \includegraphics[width=0.45\linewidth]{Figures/final/sec_model/bce/2006-05-06_RSAT_20_3.jpg}};
		\zoombox[color code=orange!70, magnification=3]{0.78,0.17}
		\zoombox[color code=cyan!70, magnification=3]{0.38,0.28}
	\end{tikzpicture}
	\caption{Glacier calving front segmentation performance of \emph{Baseline line} approach~\cite{zhang2019automatically} in 
	\subref{zoom1-image}, \subref{zoom1-zoom} versus our proposed method using the second U-Net in \subref{zoom2-image}, \subref{zoom2-zoom}. Green color represents the ground truth label, red shows the prediction, and yellow illustrates the correctly predicted pixels, \ie the overlap between the prediction and the ground truth.}
	\label{fig:zhang_vs_thres}
\end{figure*}

\section{Conclusion}\label{sec:conclusion}
This study presents a novel deep learning-based method for automatic delineation of glacier calving fronts. The main challenge in this application is the severe class-imbalance since the region of interest contains a very small portion of the pixels in the SAR images. In this work, we proposed to first regress the distance of each pixel to the calving front line. For this task, we optimized a convolutional neural network with a U-Net architecture, specially designed for glacier front segmentation.
The resulting distance map was used to localize and then generate the predicted glacier calving front line. We proposed three different approaches as a post-processing step, namely Statistical Thresholding, CRF, and Second U-Net. It has been shown that our methodology significantly outperforms the current state-of-the-art model. The Second U-Net has exhibited the best segmentation performance, which on average improves the dice coefficient 
of the best state-of-the-art model by about \SI{21}{\percent}.
The accuracy presented by our proposed method allows glaciologists to exploit it for glaciers' change detection and automatically monitor their evolution reliably.


\bibliographystyle{ieeetr}
\bibliography{References}

\end{document}